\documentclass[]{spie}  
\usepackage{amsmath}
\usepackage{algorithm,algorithmic}
   \usepackage[pdftex]{graphicx}
  \graphicspath{{Figures/}}

  \usepackage{multicol, blindtext}
\usepackage{amsmath}
\usepackage{enumerate}

\newcommand{\qed}{\nobreak \ifvmode \relax \else
      \ifdim\lastskip<1.5em \hskip-\lastskip
      \hskip1.5em plus0em minus0.5em \fi \nobreak
      \vrule height0.75em width0.5em depth0.25em\fi}

\title{Kronecker PCA Based Spatio-Temporal Modeling of Video for Dismount Classification}

\author{Kristjan H. Greenewald\supit{a} and Alfred O. Hero III\supit{a}
\skiplinehalf
\supit{a}Electrical Engineering and Computer Science, University of Michigan, Ann Arbor, MI, USA
}

\authorinfo{Further author information: (Send correspondence to Kristjan H. Greenewald at greenewk [at] umich [dot] edu)}

\begin{document}
  \maketitle

\begin{abstract}

We consider the application of KronPCA spatio-temporal modeling techniques \cite{greenewaldSSP2014,greenewaldArxiv} to the extraction of spatio-temporal features for video dismount classification. KronPCA performs a low-rank type of dimensionality reduction that is adapted to spatio-temporal data and is characterized by the $T$ frame multiframe mean $\mu$ and covariance $\mathbf{\Sigma}$ of $p$ spatial features. For further regularization and improved inverse estimation, we also use the diagonally corrected KronPCA shrinkage methods we presented in \cite{greenewaldSSP2014}. We apply this very general method to the modeling of the multivariate temporal behavior of HOG features extracted from pedestrian bounding boxes in video, with gender classification in a challenging dataset chosen as a specific application. The learned covariances for each class are used to extract spatiotemporal features which are then classified, achieving competitive classification performance. 

\end{abstract}



\section{Introduction}
\label{S:Intro}


Accurate classification of various human characteristics such as gender, gait, age, etc. in video is a crucial part of surveillance video scene understanding \cite{yu2009study}. In this work, we propose learning spatio-temporal behavior models and applying them to classifying human attributes that manifest themselves both in appearance and movement patterns. As a specific application we will consider gender classification on a challenging low-resolution surveillance video dataset.


Our approach is to first detect and track the humans in each video. At each frame, we extract spatial features from each human's area in the image.
We then propose to learn models of the spatiotemporal behavior of the features. These models are then used to create relevant spatio-temporal features with which to classify the test videos.


The spatio-temporal behavior is modeled using the spatio-temporal mean and covariance \cite{greenewaldSSP2014,greenewaldArxiv}, since the number of training samples is small relative to the number of features and we wish to avoid imposing graphical structures a priori. Our goal is thus to learn correlations between variables both within a time instant (frame) and across time using as few training samples as possible. The covariance for multivariate temporal processes manifests itself as multiframe covariance \cite{greenewaldSSP2014,greenewaldArxiv}. Let $\mathbf X$ be a $p \times T$ matrix with entries $\tilde{x}(m,t)$ denoting samples of a space-time random process defined over a $p$-grid of space samples $m\in \{1,\ldots, p\}$ and a $T$-grid of time samples $t\in \{1,\ldots, T\}$. Let ${x}={\mathrm{vec}}({\mathbf{X}})$ denote the $pT$ column vector obtained by lexicographical reordering. Define the $pT \times pT$ spatiotemporal covariance matrix
\begin{equation}
\mathbf \Sigma=\mathrm{Cov}[{x}].
\end{equation}

In this application, we must perform large $p$ and small $n$ covariance estimation, that is, we are in the high dimensional regime where the number of variables exceeds the number of training samples available to learn the covariance. For $p\geq n$, the standard sample covariance matrix (SCM) is the maximum likelihood covariance estimator. When $n$ is on the order of $p$ or smaller, however, it is well known that the SCM has a very undesirable poorly conditioned eigenstructure which results in the poor conditioning of the  inverse of the SCM, leading to poor estimates of the inverse covariance ($\mathbf{\Sigma}^{-1}$), which is needed for classification. 

One approach for addressing this problem is shrinkage estimation, which uses a weighted average of the sample covariance and a deterministic covariance, often chosen as a diagonal matrix called the shrinkage target. This can significantly improve the accuracy of the inverse of the estimate due to the improved eigenspectrum. 
In this work, we use shrinkage estimators of covariance matrices having spatiotemporal structure \cite{greenewaldSSP2014}. Estimation of spatio-temporal covariance matrices based on reducing the number of parameters via a truncated sum of Kronecker products representation, which we call KronPCA, is discussed in \cite{tsiliArxiv,greenewaldArxiv,greenewaldSSP2014}, and 
asymptotic performance analysis \cite{tsiliArxiv} predicts significant gains in estimator MSE in the high dimensional regime as $n,p \rightarrow \infty$. 



We then present a method for using the learned spatiotemporal covariances for classification, and compare it to other standard classification methods such as the SVM.

Gender classification in video is extensively covered in the literature \cite{yu2009study,hu2011gait}. Of those based on whole-body classification (e.g. as opposed to face-based classification), some such as those referenced in \cite{yu2009study} are based on various spatial features only, whereas others propose a variety of features based off of silhouettes \cite{hu2011gait,nixon2006automatic} to capture spatio-temporal behaviors such as gait (\cite{hu2011gait,nixon2006automatic} and citations). It has been found that both gait and appearance are highly indicative of gender \cite{yu2009study}, and thus aids classification significantly, with gait being especially useful when the subjects are heavily clothed and/or the video is low resolution \cite{yu2009study,hu2011gait}. Our goal is thus to reduce the potential loss of information that results from predefined feature extraction by automating the spatio-temporal modeling process while maintaining low training sample requirements, i.e. avoiding the curse of dimensionality.

The rest of this paper is organized as follows: in Section \ref{S:Kron}, we review block Toeplitz DC-KronPCA \cite{greenewaldSSP2014}. Our methods \cite{greenewaldSSP2014} for standard shrinkage of the KronPCA estimate are described in Section \ref{S:Shrink}, and our feature extraction and classification approaches are described in Section \ref{Sec:FeatClass}. The dataset we use is described in Section \ref{Sec:Data}, and gender classification results and relevant feature analysis are given in Section \ref{Sec:Result}. Our conclusions are presented in Section \ref{Sec:Conc}.

\section{Block Toeplitz Kronecker PCA}
\label{S:Kron}

In this section, we consider the regularization of the sample covariance via decomposition into a block Teoplitz sum of (space vs. time) Kronecker products representation (KronPCA) \cite{greenewaldSSP2014,tsiliArxiv,greenewaldArxiv}. 
Following \cite{greenewaldSSP2014,greenewaldArxiv,tsiliArxiv,kamm2000optimal}, the DC-KronPCA model is defined as
\begin{equation}
\label{SumApprox}
\mathbf{{\Sigma}} = \left(\sum\nolimits_{i = 1}^{r}\mathbf{T}_i \otimes \mathbf{S}_i\right) + \mathbf{I} \otimes \mathbf{U},
\end{equation}
where $\mathbf T_i$ are $T\times T$ Toeplitz matrices (the temporal Kronecker factors), $\mathbf{S}_i$ are $p\times p$ matrices (the spatial Kronecker factors), $\mathbf{U}$ is a $p\times p$ diagonal matrix.

Following the approach of \cite{tsiliArxiv}, we propose to fit the model \eqref{SumApprox} to the sample covariance matrix ${\mathbf \Sigma}_{SCM}=n^{-1}\sum_{i=1}^n (x_i-\overline{x}) (x_i-\overline{x})^T$, where $\overline{x}$ is the sample mean, and $n$ is the number of samples of the space time process $\mathbf X$. The estimation of the parameters $\mathbf T_i$, $\mathbf S_i$ and $\mathbf U$ in \eqref{SumApprox} is performed by minimizing the following objective function
\begin{equation}
\|\mathbf R-\hat{\mathbf R}\|_F^2+\beta\|\hat{\mathbf R} - \mathcal{R}(\mathbf{I}\otimes \mathbf{U})\|_*,
\end{equation}
where $\mathbf R=\mathcal{R}(\mathbf {\Sigma}_{SCM})$ and $\mathcal{R}$  denotes the permutation rearrangement operator defined in \cite{tsiliArxiv,werner2008estimation} which maps $pT\times pT$ matrices to $T^2 \times p^2$ matrices. The objective function is minimized over all $T^2 \times p^2$ matrices $\hat{\mathbf R}$ that satisfy the constraint that $\mathcal { R}^{-1}(\hat{\mathbf{ R}}) = \hat{\mathbf{\Sigma}}$  is of the form \eqref{SumApprox}.



This estimation procedure is equivalent \cite{greenewaldArxiv} to setting $\hat{\mathbf{\Sigma}} = \mathbf{I} \otimes \mathbf{U} + \mathcal{R}^{-1}(\hat{\mathbf{R}})$ where $\mathcal R^{-1}$ is the depermutation operator (inverse of $\mathcal R$) and $\hat{\mathbf{R}}$ is found by solving
\begin{align}
\min_{\hat{\mathbf{R}}} || \mathbf{M} \circ(\mathbf{R} - \hat{\mathbf{R}}) ||_F^2 + \beta \| \hat{\mathbf{R}} \|_*,\quad
\hat{\mathbf{R}} = \sum_{i = 1}^r t_i s_i^T, \:
s.t. \: \mathbf{T}_i \: \mathrm{Toeplitz}, \: \forall i,
\end{align}
where the $t_i,s_i$ are $\mathbf{T}_i, \mathbf{S}_i$ permuted as in \cite{werner2008estimation,greenewaldArxiv}, $\mathbf{M}$ is a matrix masking out the elements corresponding to the covariance diagonal, and $\circ$ denotes the elementwise product. The minimizing $\mathbf{U}$ is trivial to compute once $\hat{\mathbf{R}}$ is obtained.

Following the method of \cite{kamm2000optimal} for incorporating the Toeplitz constraint, the optimization can be shown to be equivalent to
\begin{align}
\label{Eq:OptProb}
\min_{{\tilde{\mathbf{R}}}} ||\tilde{\mathbf{M}} \circ(\mathbf{B} - {\tilde{\mathbf{R}}}) ||_F^2 + \beta \| \tilde{\mathbf{R}} \|_*
\end{align}
where $\tilde{\mathbf{R}} = \mathcal{P}(\hat{\mathbf{R}})$, $\tilde{\mathbf{M}} = \textrm{sign}(\mathcal{P}(\mathbf{M}))$, and $\mathbf{B} = \mathcal{P}(\mathbf{R})$. The operator $\mathcal{P}$ (from $T^2\times p^2$ to $(2T-1) \times p^2$ matrices) is defined as $\tilde{\mathbf{A}} = \mathcal{P}(\mathbf{A})$ such that
\begin{equation}
\tilde{A}_{j+T} = \frac{1}{\sqrt{T - |j|}}\sum_{k \in \mathcal{K}(j)}A_k,  \quad \forall j \in [-T+1, \: T-1],
\end{equation}
where $A_j$ is the $j$th row of $\mathbf{A}$.

This well-studied optimization problem (nuclear norm penalized low rank matrix approximation with missing entries) is considered in \cite{mazumder2010spectral}, where it is shown to be convex. 
The block Toeplitz diagonally corrected covariance estimate is given by
\begin{align}
\label{Eq:KronPCA}
\hat{\mathbf{\Sigma}} = \mathbf{I} \otimes \mathbf{U}+\mathcal R^{-1}\left(\mathcal{P}^{*} \left( \tilde{\mathbf{R}} \right)\right),
\end{align}
where $\tilde{\mathbf{R}}$ is the minimizer of \eqref{Eq:OptProb} and $\mathcal P^{*}$ is defined by $\mathbf{A} = \mathcal{P}^{*}(\tilde{\mathbf{A}})$ where
\begin{equation}
{A}_{k} = \frac{1}{\sqrt{T - |j|}}\tilde{A}_{j+T},  \: \forall k \in \mathcal{K}(j),\:\forall j \in [-T+1, \: T-1].
\end{equation}
Additions to the diagonal matrix $\mathbf{U}$ are discussed in the next section.
\section{Kronecker PCA Shrinkage Estimation}
\label{S:Shrink}

Diagonal shrinkage shrinks the sample covariance towards a scaled identity matrix. This improves the conditioning of the estimate, which makes the inverse of the estimate more stable.
\begin{equation}
\hat{\mathbf{\Sigma}} = (1-\hat{\rho})\hat{\mathbf{\Sigma}}_{kron} + \hat{\rho}\mathbf{F},
\end{equation}
where $\mathbf{F} = \frac{\mathrm{trace}(\hat{\mathbf{\Sigma}}_{kron})}{pT}\mathbf{I}$ and $\hat{\mathbf{\Sigma}}_{kron}$ is the DC-KronPCA estimate of the covariance. 

%

It remains to determine the amount of shrinkage, i.e. $\hat{\rho}$. We use the Ledoit-Wolf (LW) \cite{ledoit2004well} solution that asymptotically (large sample size $n$)
minimizes Frobenius estimation error when $\hat{\mathbf{\Sigma}}_{kron}$ is the SCM (i.e. full separation rank estimate). Due to the lower variance of the DC-KronPCA covariance estimate relative to the SCM, the amount of shrinkage is expected to be overestimated. We call this method DC-KronPCA-LW.

\section{Dataset}
\label{Sec:Data}

\begin{figure}[htb]
\centering
\includegraphics[width=4.85in]{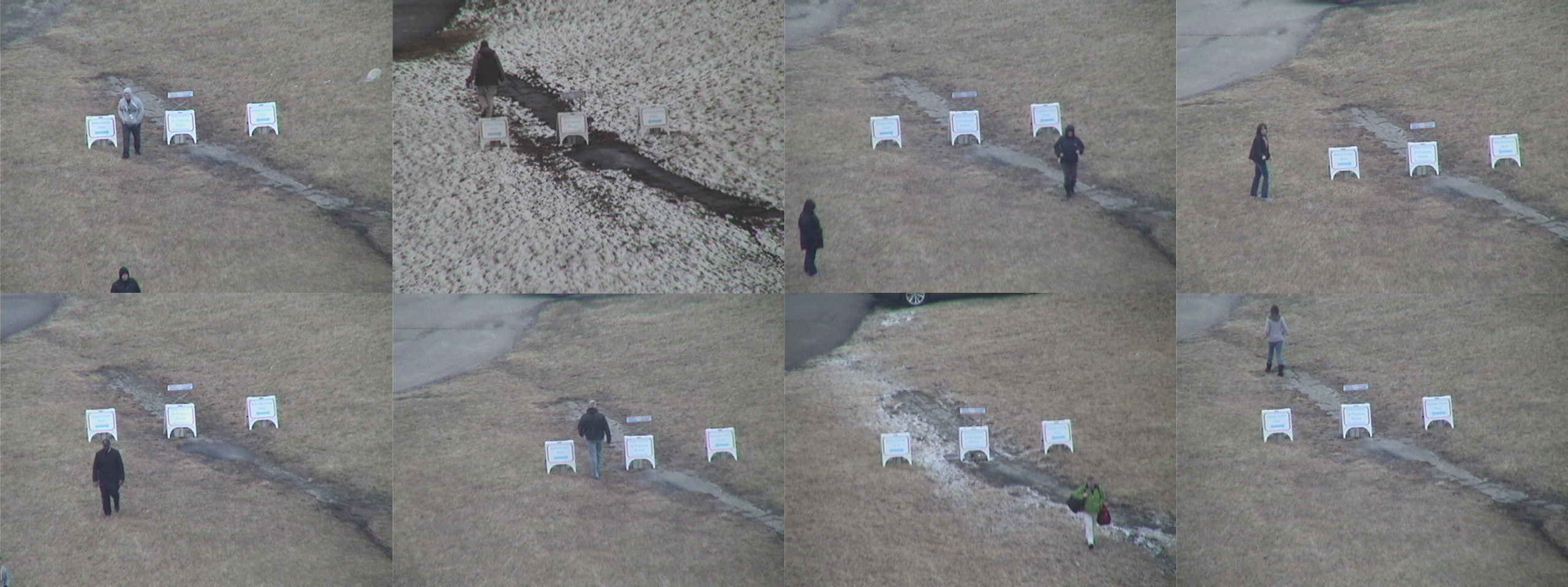}
\caption{Example video frames from SWAG-1 gender recognition dataset. Male subjects are on the left and female on the right. Note the low resolution and the variations in weather, pose, and clothing.}
\label{Fig:Frames}
\end{figure}

\begin{figure}[htb]
\centering
\includegraphics[width=2.5in]{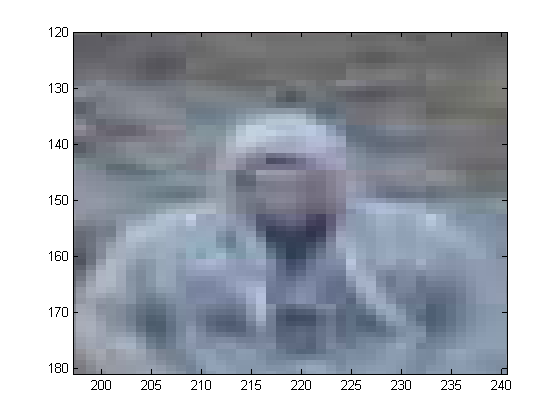}
\caption{Example zoomed in view of a face in the dataset. Note extremely low resolution.}
\label{Fig:Face}
\end{figure}

We test our methods on the SWAG-1 gender recognition dataset released by AFRL \cite{mccoppin2013electro}. This dataset consists of videos of pedestrians walking through a grassy field, imaged at a long distance using a low resolution (see Figure \ref{Fig:Face}), very low frame rate staring surveillance camera. The unstaged truthed dataset was collected over a long period, resulting in a wide variety of weather conditions such as rain, snow, fog, and sun, and a wide variety of (often heavy) clothing. There are 89 videos each for both male and female front and back views (356 total). In this work random subsets (evenly divided by gender) of the videos are used for training, and the remainder for testing, with the performance averaged over multiple Monte Carlo trials. Some example frames demonstrating the dataset variability are shown in Figure \ref{Fig:Frames}.

\section{Feature Extraction}
\label{Sec:FeatClass}

We choose to use Histogram of Oriented Gradients (HOG) based spatial features, due to their attractive invariance properties, spatial arrangement, and successful use in object detection/recognition. For each video frame, we use the Felzenswalb deformable part model HOG detector \cite{felz} to detect the humans and draw bounding boxes. Due to the relatively uniform background and relative lack of clutter, the detector performs very well. A standard tracking approach is used to connect the detections into tracks, with interpolations for (rare) missed detections. We resize the bounding boxes as needed to achieve uniformity, so that the number and position of the HOG features is invariant. Finally, for each bounding box we compute HOG features (total spatial dimension 1860). This gives a temporal sequence of features for each human track.


Due to the limited training sample size, estimation of the full covariance with sufficient accuracy for classification limits the usable number of HOG features. We thus use a dyadic blockwise approach, which involves successively splitting the HOG feature array along different spatial dimensions to create levels of nested blocks (groups of features) for which the covariances are learned independently. This in addition allows for improved robustness by generating multiple likelihood features that can be used, thus reducing any negative impacts of using a Gaussian likelihood.


For classification, we first use labeled training data to learn multiframe feature means $\mu_{k,j}$ and covariances $\mathbf{\Sigma}_{k,j}$ for each gender $k$ and each block $j$, using the learning methods discussed above. We then classify the testing data examples $x$. For each test instance, the Gaussian log likelihood ratio (LLR) across gender is computed for each block using the appropriate learned means and covariances. The block LLRs are then combined in a linear combination with positive weights learned using iteratively thresholded logistic regression. We call this approach ``KronPCA logistic LLR". Advantages of this approach are the ability to adaptively ``select" the appropriate block size for best empirical performance, reduced need for training as opposed to other approaches such as the SVM, and intuition relating to the fact that LLRs from independent observations add to form an overall LLR. As baselines we consider the SVM using the multiframe HOG features, and the standard quadratic classifier (``KronPCA overall LLR") based off the learned global means and covariances (i.e. no partition).

\section{Results}
\label{Sec:Result}
\begin{figure}[htb]
\centering
\includegraphics[width=3.5in]{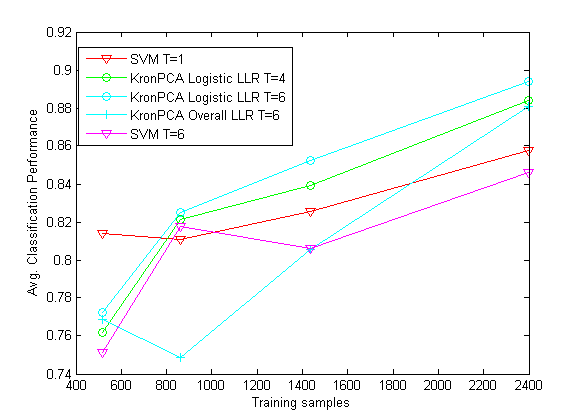}
\caption{Average correct classification rate using random training/testing data partition. Note the superiority of the KronPCA logistic multilevel LLR (loglikelihood ratio) method to the SVM and the KronPCA partition free LLR (KronPCA Overall LLR), especially for more training examples. Also note the gains achieved using larger $T$ (length of multiframe window) with the covariance methods, whereas the SVM loses performance.}
\label{Fig:Perf}
\end{figure}

Over several Monte Carlo trials, we randomly divided the dataset into $n$ frames of training tracks and $3600$ frames of testing tracks, equally divided between male and female such that the sets of videos are disjoint. We then extracted the multiframe HOG features and trained our DC-KronPCA-LW logistic LLR, the DC-KronPCA-LW overall LLR method, and the SVM. Finally, the trackwise classification performance was evaluated on the testing data. Note that the only tuning parameters are those related to the HOG features, the number of Kronecker factors used ($\beta$ chosen so that $r=2$), the number of levels in the multilevel decomposition (for logistic LLR) (held constant at 4), the length of the multiframe window $T$, and the number of training examples $n$. Results for each classifier and different values for $T$ as a function of $n$ are shown in Figure \ref{Fig:Perf} and Table \ref{Table}. Note that the logistic (multilevel) LLR outperforms the overall LLR both overall and in terms of robustness, and the SVM particularly when more training examples are available. Temporal information (used when $T>1$) improves performance as desired using our classifiers, but does not when using the SVM, indicating that the temporal regularization we employ better overcomes the curse of dimensionality stemming from the longer multiframe feature vector. For comparison, note that when the RGB pixel values in the bounding boxes are used as the only features, neither classifier exceeds coin flip rates at 2400 training samples.

\begin{figure}[htb]
\centering
\includegraphics[width=4.5in]{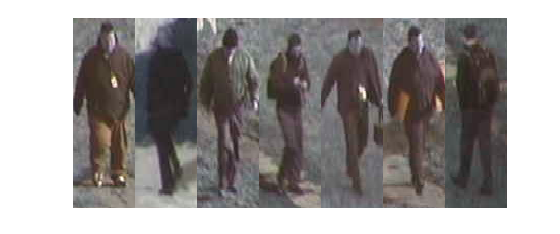}\\
\includegraphics[width=4.5in]{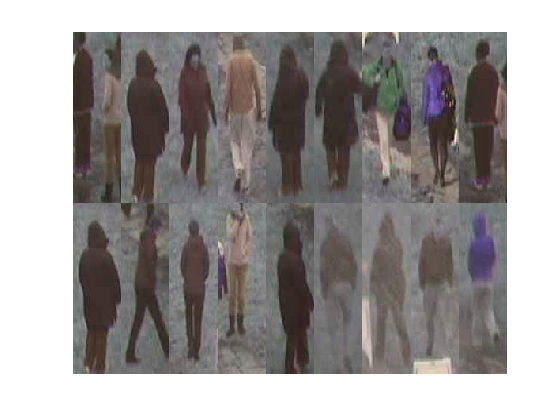}
\caption{Bounding boxes selected from each misclassified track based on using a random training set. Top: Male classified as female. Bottom: Female classified as male. Note various apparent causes of misclassification such as heavy coats, baggage, neighbor interference, heavy weather, and possibly tracks incorrectly truthed as female in the dataset.}
\label{Fig:BExamp}
\end{figure}

\begin{figure}[htb]
\centering
\includegraphics[width=4.5in]{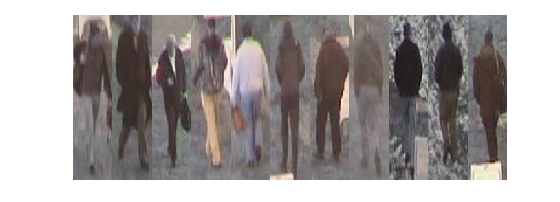}\\
\includegraphics[width=4.5in]{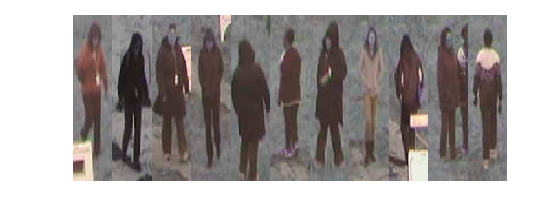}
\caption{Bounding boxes selected from example correctly classified tracks based on using a random training set. Top: Male correctly classified. Bottom: Female correctly classified. Note successes in situations including heavy weather, neighbor interference, heavy clothing/baggage, etc.}
\label{Fig:GExamp}
\end{figure}

Figure \ref{Fig:BExamp} and Figure \ref{Fig:GExamp} shows example frames from tracks that were correctly and incorrectly classified using the multilevel LLR classifier trained on a random subset of the data.

\begin{figure}[htb]
\centering
\includegraphics[width=3.4in]{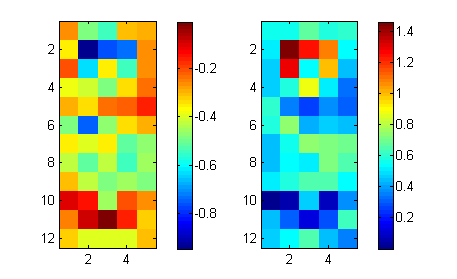}\includegraphics[width=1.6in]{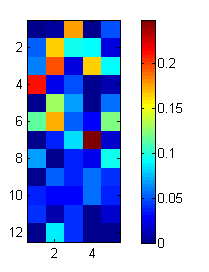}
\caption{Relevance of individual HOG features to multiframe SVM classification performance (change in correct classification \%), averaged over random data partitions). Images are of the HOG feature locations in the standard bounding boxes. Left: Performance change as a function of which HOG feature is removed. Middle: Performance change as a function of which HOG feature is added to half of the features. Right: Feature weight magnitudes learned by sparse logistic regression on the HOG features. For each of these plots, note the particular relevance of the head and leg areas, and the edge areas in the logistic regression results. }
\label{Fig:Feats}
\end{figure}

Figure \ref{Fig:Feats} shows results relating to the relevance of the spatial HOG features, specifically, the step-up and step-down performances for each feature. While no features appear to be crucial for classification, note the relative uselessness of the background areas (as expected) and the relative importance of the upper central areas corresponding to the head and shoulders area, which fits with intuition regarding gender differences in physical size, face shape, and hair.

\begin{table}[htb]
\centering
\begin{tabular}{|r|c|c|c|}
  \hline
   & $T=1$ & $T=4$ & $T=6$ \\ \hline
  KronPCA Logistic LLR & 82.1 & 88.1 & 89.4 \\ \hline
  KronPCA Overall LLR & 70.0 & 87.7 & 88.0 \\ \hline
  SVM & 85.8 & 84.6 & 84.4 \\
  \hline
\end{tabular}

\caption{Average performance (\%) for 2400 training examples. Note the superiority of KronPCA Logistic LLR for $T>1$ (i.e. when KronPCA is applicable).}
\label{Table}

\end{table}


\section{Conclusion}
\label{Sec:Conc}
In this work we considered the application of high dimensional KronPCA spatiotemporal covariance learning techniques to the modeling of the temporal behavior of spatial features in video. As an application we considered the classification of pedestrian gender in a challenging video dataset. This nonparametric modeling approach allowed the incorporation of both appearance and temporal characteristics such as gait into the features. It was found that the addition of temporal information aided classification significantly, and that our methods consistently outperform the baseline classifiers.

\acknowledgments     

This research was partially supported by ARO under grant W911NF-11-1-0391 and by AFRL under grant FA8650-07-D-1220-0006. The authors also wish to thank Mr. Edmund Zelnio for proposing the application and dataset.


\bibliography{CAMSAP_bib}   
\bibliographystyle{spiebib}   

\end{document}